\documentclass[11pt,a4paper]{article}
\usepackage[hyperref]{emnlp2020}
\usepackage{times}
\usepackage{latexsym}

\usepackage{url}
\usepackage{hyperref}

\makeatletter
\newcommand{\printfnsymbol}[1]{%
  \textsuperscript{\@fnsymbol{#1}}%
}
\makeatother

\usepackage{microtype}
\usepackage{graphicx}
\usepackage{amssymb}

\aclfinalcopy % Uncomment this line for the final submission
\title{TATL at W-NUT 2020 Task 2: A Transformer-based Baseline System for Identification of Informative COVID-19 English Tweets}

\author{Anh Tuan Nguyen \\
NVIDIA, Santa Clara, USA\\
\tt{\normalsize {tuananhn@nvidia.com}}
}

\date{}

\begin{document}
\maketitle
\begin{abstract}
As the COVID-19 outbreak continues to spread throughout the world, more and more information about the pandemic has been shared publicly on social media. For example, there are a huge number of COVID-19 English Tweets daily on Twitter. However, the majority of those Tweets are uninformative, and hence it is important to be able to automatically select only the informative ones for downstream applications. In this short paper, we present our participation in the W-NUT 2020 Shared Task 2: Identification of Informative COVID-19 English Tweets. Inspired by the recent advances in pretrained Transformer language models, we propose a simple yet effective baseline for the task. Despite its simplicity, our proposed approach shows very competitive results in the leaderboard as we ranked 8 over 56 teams participated in total.
\end{abstract}

\section{Introduction}
The COVID-19 pandemic has been spreading rapidly across the globe and has infected more than 20 millions men and women. As a result, more and more people have been sharing a wide variety of information related to COVID-19 publicly on social media. For example, there are a huge number of COVID-19 English Tweets daily on Twitter. However, the majority of those Tweets are uninformative and do not contain useful information, therefore, systems which can automatically filter out uninformative tweets are needed by the community. Tweets are generally different from traditional written-text such as Wikipedia or news articles due to its short length and informal use of words and grammars (e.g abbreviations, hashtags, marker). These special characteristics of Tweets may pose a challenge for many NLP techniques that focus solely on formally written texts.

In this paper, we present our participation in the W-NUT 2020
 Shared Task 2: Identification of Informative COVID-19 English Tweets \cite{covid19tweet}. Inspired by the recent success of Transformer-based pre-trained language models in many NLP tasks \cite{devlinetal2019bert,laietal2019gated,Chen2019BERTFJ,phobert,lai2020simple}, we propose a simple yet effective baseline for the task. Despite its simplicity, our proposed approach shows very competitive results.
 
 In the following sections, we first describe the task definitions in Section \ref{sec:overview} and proposed methods in Section \ref{sec:method}. We then describe the experiments and their results in Section \ref{sec:experiments}. Finally, in Section \ref{sec:conclusion}, we conclude this work and discuss potential future research directions.

\section{Task Definitions}\label{sec:overview}
The goal of Shared task 2 is to identify whether a COVID 19 English Tweet is informative or not. Such informative Tweet provides information about recovered, suspected, confirmed and death cases as well as location and history of each case. The dataset introduced in this Shared task consists of 10K COVID 19 English Tweets. Dataset statistics can be found in Table \ref{tab:data}

\begin{table}[!t]
    \centering
    \begin{tabular}{l|l|l|l}
    \hline
    \textbf {category} & \textbf{\#training} & \textbf{\#valid} & \textbf{\#test} \\
    \hline
    informative & 3303 & 472 & 944 \\
    uninformative & 3697 & 528 & 1056 \\
    \hline
    \end{tabular}
    \caption{Statistics of Shared task 2 dataset. ``\#training'', ``\#valid''  and  ``\#test'' denote the size of the training, validation and test sets, listed by categories, respectively.}
    \label{tab:data}
\end{table}

\section{Method}\label{sec:method}
\begin{figure*}[!t]
\centering
\includegraphics[width=\textwidth]{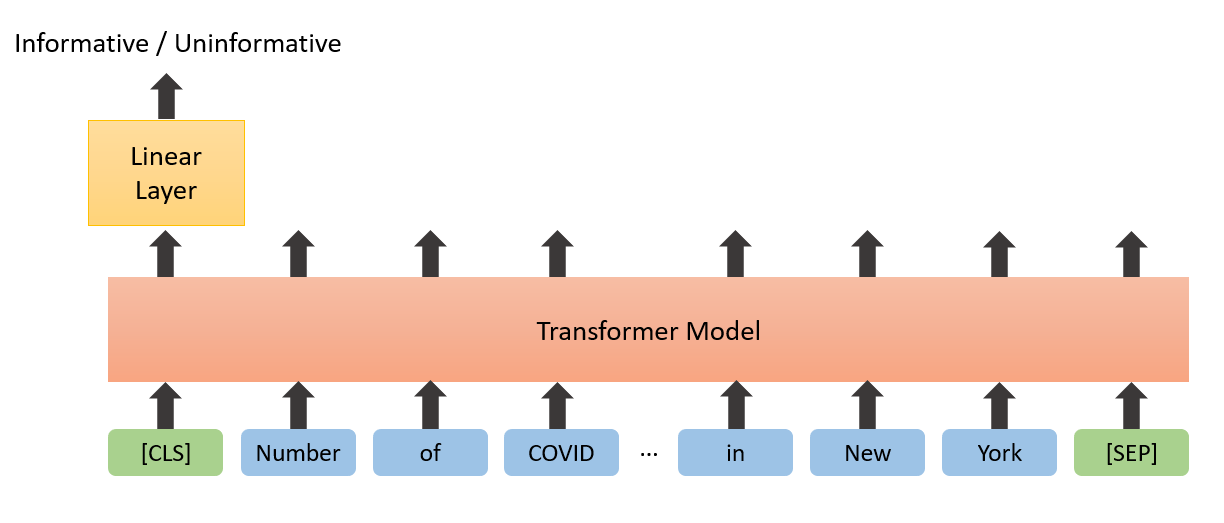}
\caption{A high level overview of our proposed model for the task.}
\label{fig:high_level}
\end{figure*}

\subsection{Baseline Model}

The task is formulated as a binary classification of Tweets into informative  or uninformative classes. Figure \ref{fig:high_level} gives a high-level overview of our proposed approach. Given 
a Tweet consisting of $n$ tokens $\textbf{x} = \{x_1, x_2, ..., x_n\}$, we first form a contextualized representation for each token using a Transformer-based encoder such as BERT \cite{devlinetal2019bert}. Following common conventions, we append special tokens to the beginning and end of the input Tweet before feeding it to the Transformer model. For example, if we use BERT, $x_1$ will be the special \texttt{[CLS]} token and $x_n$ will be the special \texttt{[SEP]} token. Let $\textbf{H} = \{\textbf{h}_1, \textbf{h}_2, ..., \textbf{h}_n\}$ denote the contextualized representations produced by the Transformer model. We then use $\textbf{h}_1$ as an aggregate representation of the original input and feed it to a linear layer to calculate the final output:
\begin{equation}
y = \sigma(\textbf{W}\textbf{h}_1 + \textbf{b})\, \in \mathbb{R} 
\end{equation}
where the transformation matrix $\textbf{W}$ and the bias term $\textbf{b}$ are model parameters. $\sigma$ denotes the sigmoid function. It squashes the score to a probability between 0 and 1. $y$ is the predicted probability of the input Tweet being informative. 

In this work, we experiment with various state-of-the-art Transformer models including BERTweet \cite{Nguyen2020BERTweetAP}, XLM-RoBERTa \cite{conneauetal2020unsupervised},
 RoBERTa \cite{RoBERTa}, and ELECTRA  \cite{Clark2020ELECTRAPT}. In the following subsections, we will briefly describe these Transformer models. 

\subsubsection{RoBERTa}
RoBERTa \cite{RoBERTa} improved over BERT \cite{devlinetal2019bert} by leveraging different training objectives which leads to more robust optimization i.e removing next sentence prediction and using dynamic masking for masked language modelling. \newcite{RoBERTa} also shows that training the language model longer and with more data hugely benefits the performance on downstream tasks.

\subsubsection{XLM-RoBERTa}
Inspired by the success of multilingual language model \cite{devlinetal2019bert, lample2019cross}, XLM-RoBERTa \citep{conneauetal2020unsupervised} significantly scaled up the amount of multilingual training data used in unsupervised MLM pre-training compares to previous work \citep{lample2019cross} and achieved state-of-the-art performance in both monolingual and cross-lingual benchmarks. 

\subsubsection{BERTweet}
BERTweet \citep{Nguyen2020BERTweetAP} is a domain-specific language model pre-trained on a large corpus of English Tweets. Similar to the success of BioBERT \citep{10.1093/bioinformatics/btz682} in BioNLP domain and the success of SciBERT \citep{Beltagy2019SciBERT} in ScientificNLP domain, BERTweet achieved state-of-the-art performance across many TweetNLP tasks, outperformed its counterparts RoBERTa \citep{RoBERTa} and XLM-RoBERTa \citep{conneauetal2020unsupervised}.

\subsubsection{ELECTRA}
ELECTRA \citep{Clark2020ELECTRAPT} proposed a new pre-training objective which is different from Masked Language Modelling \citep{devlinetal2019bert, RoBERTa}. Instead of masking input tokens, ELECTRA corrupts the tokens using a small generator network to produces distribution over tokens, while the discriminator tries to guess which tokens are actually corrupted by the generator. ELECTRA achieved state-of-the-art results across many tasks in the GLUE benchmark \citep{DBLP:conf/iclr/WangSMHLB19} while using much less compute resources compared to other pre-training methods \citep{devlinetal2019bert, RoBERTa}.

\subsection{Ensemble Learning}
To further boost the performance of our baseline models, we leverage ensemble learning technique. We performed ensemble learning over all of the Transformer models mentioned in the previous section and employed two different ensemble schemes, namely Unweighted Averaging and Majority Voting. 
\subsubsection{Unweighted Averaging}
In this approach, the final prediction is estimated from the unweighted average of the posterior probability from all of our models. Thus, the final prediction is given by: \\
\begin{equation}
p = \arg\max_{c} \frac{1}{M} \sum_{n=1}^{M} p_i, \hspace{0.25cm} p_i \in \mathbb{R ^ {C}}
\end{equation}
\\
where \(C\) is the number of classed, \(M\) is the number of models, and \(p_i\) is the probability vector computed using the softmax function of model \(i\).

\subsubsection{Majority Voting}
Majority Voting counts the votes of all the models and select the class with most votes as prediction. Formally, the final prediction is given by:
\\
\begin{equation}
v_c = \sum_{n=1}^{M} F_i(c), \hspace{0.25cm} p = \arg\max_{c} v_c
\end{equation}
\\
where \(v_c\) denotes the votes of class \(c\) from all different models, \(F_i\) is the binary decision of model \(i\), which is either 0 or 1.

\section{Experiments}\label{sec:experiments}
\subsection{Finetuning}
To fine-tune our baseline models, we employ \texttt{transformers} library \citep{Wolf2019HuggingFacesTS}. We use AdamW optimizer \citep{loshchilov2018decoupled} with a fixed batch size of 32 and learning rates in the set $\{1e-5, 2e-5, 5e-5\}$. We fine-tune the models for 30 epochs and select the best checkpoint based on performance of the model on the validation set.

\subsection{Performance of our baselines}
Table \ref{tab:individual_results_val_set} shows the overall results on the validation set. The large version of RoBERTa achieves the highest F1 score on the validation set (compared to other individual models). To our surprise, we find that BERTweet does not outperform the base version of RoBERTa on the validation set, even though BERTweet was trained on English Tweets using the same training procedure of RoBERTa. Finally, XLM-RoBERTa achieves lower F1 score than both RoBERTa and ELECTRA, suggesting that using a multilingual pretrained language models may not improve the performance since the shared task is mainly about English Tweets. We also evaluate the performance of our ensemble models. The results show that ensemble learning improves the F1 score compare to each individual model and Unweighted Averaging perform better than Majority Voting on the validation set. We also submitted the predictions of both ensemble scheme to the competition and final results on the leaderboard are shown in table \ref{tab:result_test_set}. We notice that Majority Voting slightly performs better than Unweighted Averaging on the hidden test set. 

\begin{table}[t]
\centering
\begin{tabular}{|l|c|}
\hline
\multicolumn{1}{|c|}{Model} & Dev F1 \\ \hline
XLM-RoBERTa (base) & 0.905 \\
XLM-RoBERTa (large) & 0.906 \\ \hline
RoBERTa (base) & 0.911 \\
RoBERTa (large) & \textbf{0.918} \\ \hline
BERTweet & 0.909 \\\hline
ELECTRA (base) & 0.907\\
ELECTRA (large) & 0.914\\ \hline
Ensemble (averaging) & \textbf{0.927} \\
Ensemble (voting) & 0.922 \\ \hline
\end{tabular}
\caption{Performance of individual models as well as ensemble models on the validation set.}
\label{tab:individual_results_val_set}
\end{table}

\begin{table}[t]
\centering
\begin{tabular}{|l|c|}
\hline
\multicolumn{1}{|c|}{Model} & Test F1 \\ \hline
Ensemble (averaging) & 0.8988 \\ \hline
Ensemble (voting) & \textbf{0.9008} \\ \hline
\end{tabular}
\caption{Performance of our system on the test set.}
\label{tab:result_test_set}
\end{table} 

\section{Conclusion}\label{sec:conclusion}
In this paper, we introduce a simple but effective approach for identifying informative COVID-19 English Tweets. Despite the simplicity of our approach, it achieves very competitive results in the leaderboard as we ranked 8 over 56 teams participated in total. In future work, we will conduct thorough error analysis and apply visualization techniques to gain more understandings of our models \cite{murugesan2019deepcompare}. Furthermore, we will also extend our approach to other languages. Finally, we will investigate the use of advanced techniques such as transfer learning, few-shot learning, and self-training to improve the performance of our system further \cite{panetal2017cross,huangetal2018zero,lai2018supervised,yoon2019compare,Xie2020SelfTrainingWN}.

\bibliographystyle{acl_natbib}
\bibliography{anthology,emnlp2020}

\end{document}